\documentclass[11pt]{article}
\usepackage{naaclhlt2009}
\usepackage{times}
\usepackage{latexsym}
\usepackage{haldefs}
\usepackage[pdftex]{graphicx}
\title{Non-Parametric Bayesian Areal Linguistics}

\newcommand{\mysection}[1]{\vspace{-3mm}\section{#1}\vspace{-2mm}}
\newcommand{\mysubsection}[1]{\vspace{-2mm}\subsection{#1}\vspace{-1mm}}
\newcommand{\mysubsubsection}[1]{\vspace{-2mm}\subsubsection{#1}\vspace{-1mm}}

\newcommand{\mycaption}[1]{\vspace{-0.4em}\caption{#1}\vspace{-0.7em}}

\author{Hal Daum\'e III \\
  School of Computing\\
  University of Utah\\
  Salt Lake City, UT 84112\\
  {\tt me@hal3.name}}

\date{}

\begin{document}
\maketitle

\begin{abstract}
  We describe a statistical model over linguistic areas and phylogeny.
  Our model recovers known areas and identifies a plausible hierarchy
  of areal features.  The use of areas improves genetic reconstruction
  of languages both qualitatively and quantitatively according to a
  variety of metrics.  We model linguistic areas by a Pitman-Yor
  process and linguistic phylogeny by Kingman's coalescent.
\end{abstract}

\mysection{Introduction} \label{sec:intro}

Why are some languages more alike than others?  This question is one
of the most central issues in historical linguistics.  Typically, one
of three answers is given~\cite{aikhenvald01book,campbell06problem}.
First, the languages may be related ``genetically.''  That is, they
may have all derived from a common ancestor language.  Second, the
similarities may be due to chance.  Some language properties are
simply more common than others, which is often attributed to be mostly
due to linguistic universals \cite{greenberg63universals}.  Third, the
languages may be related \emph{areally}.  Languages that occupy the
same geographic area often exhibit similar characteristics, not due to
genetic relatedness, but due to sharing.  Regions (and the languages
contained within them) that exhibit sharing are called
\emph{linguistic areas} and the features that are shared are called
\emph{areal features}.

Much is not understood or agreed upon in the field of areal
linguistics.  Different linguists favor different defintions of what
it means to be a linguistic area (are two languages sufficient to
describe an area or do you need three
\cite{thomason01contact,katz75borrow}?), what areal features are (is there a
linear ordering of ``borrowability''
\cite{katz75borrow,curnow01borrowability} or is that too
prescriptive?), and what causes sharing to take place (does social
status or number of speakers play a role \cite{thomason01contact}?).

In this paper, we attempt to provide a \emph{statistical} answer to
some of these questions.  In particular, we develop a Bayesian model
of typology that allows for, but does not force, the existence of
linguistic areas.  Our model also allows for, but does not force,
preference for some feature to be shared areally.  When applied to a
large typological database of linguistic features \cite{wals}, we find
that it discovers linguistic areas that are well documented in the
literature (see Campbell~\shortcite{campbell05areal} for an overview),
and a small preference for certain features to be shared areally.
This latter agrees, to a lesser degree, with some of the published
hierarchies of borrowability \cite{curnow01borrowability}.  Finally,
we show that reconstructing language family trees is significantly
aided by knowledge of areal features.  We note that Warnow et
al.~\shortcite{warnow05borrowing} have independently proposed a model
for phonological change in Indo-European (based on the Dyen dataset
\cite{dyen92indoeuropean}) that includes notions of borrowing.  Our
model is different in that we (a) base our model on typological
features rather than just lexical patterns and (b) we explicitly
represent language areas, not just one-time borrowing phenomena.

\mysection{Background} \label{sec:bg}

We describe (in Section~\ref{sec:model}) a non-parametric,
hierarchical Bayesian model for finding linguistic areas and areal
features.  In this section, we provide necessary background---both
linguistic and statistical---for understanding our model.

\mysubsection{Areal Linguistics} \label{sec:areal}

Areal effects on linguistic typology have been studied since, at
least, the late 1920s by Trubetzkoy, though the idea of tracing family
trees for languages goes back to the mid 1800s and the comparative
study of historical linguistics dates back, perhaps to Giraldus
Cambrenis in 1194 \cite{campbell-jones}.  A recent article provides a
short introduction to both the issues that surround areal linguistics,
as well as an enumeration of many of the known language areas
\cite{campbell05areal}.  A fairly wide, modern treatment of the issues
surrounding areal diffusion is also given by essays in a recent book
edited by Aikhenvald and Dixon~\shortcite{aikhenvald01book}.  The
essays in this book provide a good introduction to the issues in the
field.  Campbell~\shortcite{campbell06problem} provides a critical
survey of these and other hypotheses relating to areal linguistics.

There are several issues which are basic to the study of areal
linguistics (these are copied almost directly from Campbell
\shortcite{campbell06problem}).  Must a linguistic area
comprise more than two languages?  Must it comprise more than one
language family?  Is a single trait sufficient to define an area?  How
``nearby'' must languages in an area be to one another?  Are some
feature more easily borrowed that others?






Despite these formal definitional issues of what constitutes a
language area and areal features, most historical linguists seem
to believe that areal effects play \emph{some} role in the change of
languages.

\mysubsubsection{Established Linguistic Areas} \label{sec:areas}

Below, we list some of the well-known linguistic areas;
Campbell~\shortcite{campbell05areal} provides are more complete
listing together with example areal features for these areas.  For
each area, we list associated languages:

\noindent
{\bf The Balkans:} Albanian, Bulgarian, Greek, Macedonian, Rumanian
  and Serbo-Croatian.  (\emph{Sometimes:} Romani and Turkish)

\noindent
{\bf South Asian:} Languages belonging to the Dravidian, Indo-Aryan,
  Munda, Tibeto-Burman families.

\noindent
{\bf Meso-America:} 
Cuitlatec,
Huave,
Mayan,
Mixe-Zoquean,
Nahua,
Otomanguean,
Tarascan,
Tequistlatecan,
Totonacan and
Xincan.

\noindent
{\bf North-west America:}
Alsea,
Chimakuan,
Coosan,
Eyak,
Haida,
Kalapuyan,
Lower Chinook,
Salishan,
Takelman,
Tlingit,
Tsimshian and
Wakashan.

\noindent
{\bf The Baltic:}
Baltic languages,
Baltic German, and
Finnic languages (especially Estonian and Livonian).  (Sometimes many
more are included, such as:
Belorussian,
Lavian,
Lithuanian,
Norwegian,
Old Prussian,
Polish,
Romani,
Russian,
Ukranian.)

\noindent
{\bf Ethiopia:}
Afar, 
Amharic,
Anyuak,
Awngi, 
Beja, 
Ge'ez,
Gumuz,
Janjero,
Kefa,
Sidamo, 
Somali,
Tigre,
Tigrinya and
Wellamo.

Needless to say, the exact definition and extent of the actual areas
is up to significant debate.  Moreover, claims have been made in favor
of many linguistic areas not defined above.  For instance, Dixon
\shortcite{dixon01australian} presents arguments for several
Australian linguistic areas and Matisoff~\shortcite{matisoff01sea}
defines a South-East Asian language area.  Finally, although ``folk
lore'' is in favor of identifying a linguistic area including English,
French and certain Norse languages (Norwegian, Swedish, Low Dutch,
High German, etc.), there are counter-arguments to this position
\cite{thomason01contact} (see especially Case Study 9.8).

\mysubsubsection{Linguistic Features} \label{sec:features}

Identifying which linguistic features are most easily shared
``areally'' is a long standing problem in contact linguistics.  Here
we briefly review some of the major claims.  Much of this overview is
adoped from the summary given by
Curnow~\shortcite{curnow01borrowability}.

Haugen~\shortcite{haugen50borrow} considers only borrowability as far
as the lexicon is concerned.  He provided evidence that nouns are the
easiest, followed by verbs, adjectives, adverbs, prepositions, etc.
Ross~\shortcite{ross88borrow} corroborates Haugen's analysis and
deepens it to cover morphology, syntax and phonology.  He proposes the
following hierarchy of borrowability (easiest items coming first):
nouns $>$ verbs $>$ adjectives $>$ syntax $>$ non-bound function words
$>$ bound morphemes $>$ phonemes.  Coming from a ``constraints''
perspective, Moravcsik~\shortcite{moravcsik78constraints} suggests
that: lexical items must be borrowed before lexical properties;
inflected words before bound morphemes; verbal items can never be
borrowed; etc.

Curnow~\shortcite{curnow01borrowability} argues that coming up with a
reasonable hierarchy of borrowability is that ``we may never be able
to develop such constraints.''  Nevertheless, he divides the space of
borrowable features into 15 categories and discusses the evidence
supporting each of these categories, including: phonetics (rare),
phonology (common), lexical (very common), interjections and discourse
markers (common), free grammatical forms (occasional), bound
grammatical forms (rare), position of morphology (rare), syntactic
frames (rare), clause-internal syntax (common), between-clause syntax
(occasional).

\mysubsection{Non-parametric Bayesian Models} \label{sec:bayes}

We treat the problem of understanding areal linguistics as a
statistical question, based on a database of typological information.
Due to the issues raised in the previous section, we do not want to
commit to the existence of a particular number of linguistic areas, or
particular sizes thereof.  (Indeed, we do not even want to commit to
the existence of \emph{any} linguistic areas.)  However, we will need
to ``unify'' the languages that fall into a linguistic area (if such a
thing exists) by means of some statistical parameter.  Such problems
have been studied under the name \emph{non-parametric models.}  The
idea behind non-parametric models is that one does not commit \emph{a
  priori} to a particularly number of parameters.  Instead, we allow
the data to dictate how many parameters there are.  In Bayesian
modeling, non-parametric distributions are typically used as
\emph{priors}; see Jordan~\shortcite{jordan05nonparametric} or
Ghahramani~\shortcite{ghahramani05nonparametric} for overviews.  In
our model, we use two different non-parametric priors: the Pitman-Yor
process (for modeling linguistic areas) and Kingman's coalescent (for
modeling linguistic phylogeny), both described below.

\mysubsubsection{The Pitman-Yor Process} \label{sec:py}

One particular example of a non-parametric prior is the Pitman-Yor
process \cite{pitman-yor97}, which can be seen as an extension to the
better-known Dirichlet process \cite{ferguson74priordistributions}.
The Pitman-Yor process can be understood as a particular example of a
Chinese Restaurant process (CRP) \cite{pitman02crp}.  The idea in all
CRPs is that there exists a restaurant with an infinite number of
tables.  Customers come into the restaurant and have to choose a table
at which to sit.

The Pitman-Yor process is described by three parameters: a base rate
$\al$, a discount parameter $d$ and a mean distribution $G_0$.  These
combine to describe a process denoted by $\PY(\al,d,G_0)$.  The
parameters $\al$ and $d$ must satisfy: $0 \leq d < 1$ and $\al > -d$.
In the CRP analogy, the model works as follows.  The first customer
comes in and sits at any table.  After $N$ customers have come in and
seated themselves (at a total of $K$ tables), the $N$th customer
arrives.  In the Pitman-Yor process, the $N$th customer sits at a new
table with probability proportional to $\al + Kd$ and sits at a
previously occupied table $k$ with probability proportional to $\#_k -
d$, where $\#_k$ is the number of customers already seated at table
$k$.  Finally, with each table $k$ we associate a parameter $\th_k$,
with each $\th_k$ drawn independently from $G_0$.  An important
property of the Pitman-Yor process is that draws from it are
\emph{exchangable}: perhaps counterintuitively, the distribution does
not care about customer order.

The Pitman-Yor process induces a power-law distribution on the number
of singleton tables (i.e., the number of tables that have only one
customer).  This can be seen by noticing two things.  In general, the
number of singleton tables grows as $\cO(\al N^d)$.  When $d=0$, we
obtain a Dirichlet process with the number of singleton tables growing
as $\cO(\al \log N)$.

\mysubsubsection{Kingman's Coalescent} \label{sec:coalescent}

Kingman's coalescent is a standard model in population genetics
describing the common genealogy (ancestral tree) of a set of
individuals \cite{Kin1982a,Kin1982b}.  In its full form it is a
distribution over the genealogy of a countable set.

Consider the genealogy of $n$ individuals alive at the present time
$t=0$.  We can trace their ancestry backwards in time to the distant
past $t\!=\!-\!\infty$.  Assume each individual has one parent (in
genetics, \emph{haploid} organisms), and therefore genealogies of
$[n]=\{1,\ldots,n\}$ form a \emph{directed forest}.  Kingman's
$n$-coalescent is simply a distribution over genealogies of $n$
individuals.  To describe the Markov process in its entirety, it is
sufficient to describe the jump process (i.e.\ the embedded,
discrete-time, Markov chain over partitions) and the distribution over
coalescent times.  In the $n$-coalescent, every pair of lineages
merges independently with rate 1, with parents chosen uniformly at
random from the set of possible parents at the previous time step.

The $n$-coalescent has some interesting statistical properties
\cite{Kin1982a,Kin1982b}.  The marginal distribution over tree
topologies is uniform and independent of the coalescent times.
Secondly, it is infinitely exchangeable: given a genealogy drawn from
an $n$-coalescent, the genealogy of any $m$ contemporary individuals
alive at time $t\!\le\!0$ embedded within the genealogy is a draw from
the $m$-coalescent. Thus, taking $n\!\rightarrow\!\infty$, there is a
distribution over genealogies of a countably infinite population for
which the marginal distribution of the genealogy of any $n$
individuals gives the $n$-coalescent.  Kingman called this \emph{the
  coalescent}.

Teh et al. \shortcite{teh07coalescent} recently described efficient
inference algorithms for Kingman's coalescent.  They applied the
coalescent to the problem of recovering linguistic phylogenies.  The
application was largely successful---at least in comparison to
alternative algorithms that use the same data-.  Unfortunately, even
in the results they present, one can see significant areal effects.
For instance, in their Figure(3a), Romanian is very near Albanian and
Bulgarian.  This is likely an areal effect: specifically, an effect
due to the Balkan langauge area.  We will revisit this issue in our
own experiments.

\mysection{A Bayesian Model for Areal Linguistics} \label{sec:model}

We will consider a data set consisting of $N$ languages and $F$
typological features.  We denote the value of feature $f$ in language
$n$ as $X_{n,f}$.  For simplicity of exposition, we will assume two
things: (1) there is no unobserved data and (2) all features are
binary.  In practice, for the data we use (described in
Section~\ref{sec:data}), neither of these is true.  However, both
extensions are straightforward.

When we construct our model, we attempt to be as neutral to the
``areal linguistics'' questions defined in Section~\ref{sec:areal} as
possible.  We allow areas with only two languages (though for brevity
we do not present them in the results).  We allow areas with only one
family (though, again, do not present them).  We are generous with our
notion of locality, allowing a radius of $1000$ kilometers (though see
Section~\ref{sec:radius} for an analysis of the effect of
radius).\footnote{An reader might worry about exchangeability:
  Our method of making language centers and locations part of the
  Pitman-Yor distribution ensures this is not an issue.  An
  alternative would be to use a location-sensitive process such as the
  kernel stick-breaking process \cite{dunson07ksbp}, though we do not
  explore that here.}  And we allow, but do not enforce trait weights.
All of this is accomplished through the construction of the model and
the choice of the model hyperparameters.

At a high-level, our model works as follows.  Values $X_{n,f}$ appear
for one of two reasons: they are either areally derived or genetically
derived.  A latent variable $Z_{n,f}$ determines this.  If it is derived
areally, then the value $X_{n,f}$ is drawn from a latent variable
corresponding to the value preferences in the langauge area to which
language $n$ belongs.  If it is derived genetically, then $X_{n,f}$ is
drawn from a variable corresponding to value preferences for the
genetic substrate to which language $n$ belongs.  The set of areas,
and the area to which a language belongs are given by yet more latent
variables.  It is this aspect of the model for which we use the
Pitman-Yor process: languages are customers, areas are tables and area
value preferences are the parameters of the tables.

\mysubsection{The formal model} \label{sec:formal}

We assume that the value a feature takes for a particular language
(i.e., the value of $X_{n,f}$) can be explained \emph{either}
genetically or areally.\footnote{As mentioned in the introduction, (at
  least) one more option is possible: chance.  We treat ``chance'' as
  noise and model it in the data generation process, not as an
  alternative ``source.''}  We denote this by a binary indicator
variable $Z_{n,f}$, where a value $1$ means ``areal'' and a value $0$
means ``genetic.''  We assume that each $Z_{n,f}$ is drawn from a
feature-specific binomial parameter $\pi_f$.  By having the parameter
feature-specific, we express the fact that some features may be more
or less likely to be shared than others.  In other words, a high value
of $\pi_f$ would mean that feature $f$ is easily shared areally, while
a low value would mean that feature $f$ is hard to share.  Each
language $n$ has a known latitude/longitude $\ell_n$.

\begin{figure*}[t]
\begin{small}
\begin{tabular}{|r@{ }c@{ }ll|}
\hline
$X_{n,f}$    &$\sim$& $\brack{\Bin(\th_{p_n,f}) & \text{if } Z_{n,f} = 0 \\
                   \Bin(\ph_{a_n,f}) & \text{if } Z_{n,f} = 1}$ 
& feature values are derived genetically or areally \\
$Z_{n,f}$   &$\sim$& $\Bin(\pi_f)$ &
feature source is a biased coin, parameterized per feature \\
$\ell_n$    &$\sim$& $\cB\hspace{-0.5mm}\textit{all}(c_{a_n}, R)$ &
language position is uniform within a ball around area center, radius $R$ \\
$\pi_f$     &$\sim$& $\Bet(1,1)$ &
bias for a feature being genetic/areal is uniform \\
$(p,\th)$   &$\sim$& $\textrm{Coalescent}(\pi_0,m_0)$ &
language hierarchy and genetic traits are drawn from a Coalescent \\
$(a,\langle \ph, c \rangle)$ &$\sim$& $\PY(\al_0, d_0, \Bet(1,1) \times \Uni)$ &
area features are drawn Beta and centers Uniformly across the globe \\
\hline
\end{tabular}
\end{small}
\mycaption{Full hierarchical Areal model; see Section~\ref{sec:formal}
  for a complete description.}
\label{fig:areal}
\end{figure*}

We further assume that there are $K$ linguistic areas, where $K$ is
treated non-parametrically by means of the Pitman-Yor process.  Note
that in our context, a linguistic area may contain \emph{only one}
language, which would technically not be allowed according to the
linguistic definition.  When a language belongs to a singleton area,
we interpret this to mean that it does not belong to any language
area.

Each language area $k$ (including the singleton areas) has a set of
$F$ associated parameters $\phi_{k,f}$, where $\phi_{k,f}$ is the
probability that feature $f$ is ``on'' in area $k$.  It also has a
``central location'' given by a longitude and latitude denoted $c_k$.
We only allow languages to belong to areas that fall within a given
radius $R$ of them (distances computed according to geodesic
distance).  This accounts for the ``geographical'' constraints on
language areas.  We denote the area to which language $n$ belongs as
$a_n$.

We assume that each language belongs to a ``family tree.''  We denote
the parent of language $n$ in the family tree by $p_n$.  We associate
with each node $i$ in the family tree and each feature $f$ a parameter
$\th_{i,f}$.  As in the areal case, $\th_{i,f}$ is the probability
that feature $f$ is on for languages that descend from node $i$ in the
family tree.  We model genetic trees by Kingman's
coalescent with binomial mutation.

Finally, we put non-informative priors on all the hyperparameters.
Written hierarchically, our model has the following shown in
Figure~\ref{fig:areal}.  There, by $(p,\th) \by
\textrm{Coalescent}(\pi_0,m_0)$, we mean that the tree and parameters
are given by a coalescent.

\mysubsection{Inference} \label{sec:inference}

Inference in our model is mostly by Gibbs sampling.  Most of the
distributions used are conjugate, so Gibbs sampling can be implemented
efficiently.  The only exceptions are: (1) the coalescent for which we
use the GreedyRate1 algorithm described by Teh et al.
\shortcite{teh07coalescent}; (2) the area centers $c$, for which we
using a Metropolis-Hastings step.  Our proposal distribution is a
Gaussian centered at the previous center, with standard deviation of
$5$.  Experimentally, this resulted in an acceptance rate of about
$50\%$.

In our implementation, we analytically integrate out $\pi$ and $\phi$
and sample only over $Z$, the coalescent tree, and the area
assignments.  In some of our experiments, we treat the family tree as
given.  In this case, we also analytically integrate out the $\th$
parameters and sample only over $Z$ and area assignments.


\mysection{Typological Data} \label{sec:data}


The database on which we perform our analysis is the \emph{World Atlas
  of Language Structures} (henceforth, WALS) \cite{wals}.  The
database contains information about $2150$ languages (sampled from
across the world).  There are $139$ typological \emph{features} in
this database.  The database is \emph{sparse}: only $16\%$ of the
possible language/feature pairs are known.  We use the version
extracted and preprocessed by Daum\'e III and
Campbell~\shortcite{daume07implication}.

In WALS, languages a grouped into 38 language families (including
Indo-European, Afro-Asiatic, Austronesian, Niger-Congo, etc.).  Each
of these language families is grouped into a number of language geni.
The Indo-European family includes ten geni, including: Germanic,
Romance, Indic and Slavic.  The Austronesian family includes seventeen
geni, including: Borneo, Oceanic, Palauan and Sundic.  Overall, there
are 275 geni represented in WALS.

We further preprocess the data as follows.  For the Indo-European
subset (hence-forth, ``IE''), we remove all languages with $\leq 10$
known features and then remove all features that appear in at most
$1/4$ of the languages.  This leads to $73$ languages and $87$
features.  For the whole-world subset, we remove languages with $\leq
25$ known features and then features that appear in at most $1/10$ of
the languages.  This leads to $349$ languages and $129$ features.

\mysection{Experiments} \label{sec:experiments}
\vspace{2mm}

\mysubsection{Identifying Language Areas} \label{sec:experiment-areas}

Our first experiment is aimed at discovering language areas.  We first
focus on the IE family, and then extend the analysis to all languages.
In both cases, we use a known family tree (for the IE experiment, we
use a tree given by the language genus structure; for the whole-world
experiment, we use a tree given by the language family structure).  We
run each experiment with five random restarts and 2000 iterations.  We
select the MAP configuration from the combination of these runs.

\begin{figure}[t]
\footnotesize
\centering
\begin{tabular}{@{}p{8cm}@{}}
\hline
{\bf (Indic)} Bhojpuri, Darai, Gujarati, Hindi, Kalami, Kashmiri, Kumauni, Nepali, Panjabi, Shekhawati, Sindhi 
{\bf (Iranian)} Ormuri, Pashto \\
\hline
{\bf (Albanian)} Albanian 
{\bf (Greek)} Greek (Modern) 
{\bf (Indic)} Romani (Kalderash) 
{\bf (Romance)} Romanian, Romansch (Scharans), Romansch (Sursilvan), Sardinian 
{\bf (Slavic)} Bulgarian, Macedonian, Serbian-Croatian, Slovak, Slovene, Sorbian \\
\hline
{\bf (Baltic)} Latvian, Lithuanian 
{\bf (Germanic)} Danish, Swedish 
{\bf (Slavic)} Polish, Russian \\
\hline
{\bf (Celtic)} Irish 
{\bf (Germanic)} English, German, Norwegian 
{\bf (Romance)} French \\
\hline
{\bf (Indic)} Prasuni, Urdu 
{\bf (Iranian)} Persian, Tajik \\
\hline
{\bf Plus 46 non-areal languages} \\
\hline
\end{tabular}
  \label{fig:ie-areas}
  \mycaption{IE areas identified.  Areas that consist of just one
genus are not listed, nor are areas with two languages.}
\end{figure}

\begin{figure}[t]
  \centering
\footnotesize
\centering
\begin{tabular}{@{}p{8cm}@{}}
\hline
{\bf (Mayan)} Huastec, Jakaltek, Mam, Tzutujil
{\bf (Mixe-Zoque)} Zoque (Copainal\'a)
{\bf (Oto-Manguean)} Mixtec (Chalcatongo), Otom\'i (Mezquital)
{\bf (Uto-Aztecan)} Nahualtl (Tetelcingo), Pipil\\
\hline
{\bf (Baltic)} Latvian, Lithuanian
{\bf (Finnic)} Estonian, Finnish
{\bf (Slavic)} Polish, Russian, Ukranian\\
\hline
{\bf (Austro-Asiatic)} Khasi
{\bf (Dravidian)} Telugu
{\bf (IE)} Bengali
{\bf (Sino-Tibetan)} Bawm, Garo, Newari (Kathmandu)\\
\hline
\end{tabular}
  \label{fig:world-areas}
  \mycaption{A small subset of the world areas identified.}
\end{figure}

In the IE experiment, the model identified the areas shown in
Figure~\ref{fig:ie-areas}.  The best area identified by our model is
the second one listed, which clearly correlates highly with the
Balkans.  There are two areas identified by our model (the first and
last) that include only Indic and Iranian languages.  While we are not
aware of previous studies of these as linguistic areas, they are not
implausible given the history of the region.  The fourth area
identified by our model corresponds roughly to the debated ``English''
area.  Our area includes the requisite French/English/German/Norwegian
group, as well as the somewhat surprising Irish.  However, in addition
to being intuitively plausible, it is not hard to find evidence in the
literature for the contact relationship between English and Irish
\cite{sommerfelt60factors}.

In the whole-world experiment, the model identified too many
linguistic areas to fit ($39$ in total that contained at least two
languages, and contained at least two language families).  In
Figure~\ref{fig:world-areas}, we depict the areas found by our model
that best correspond to the areas described in
Section~\ref{sec:areas}.  We acknowledge that this gives a warped
sense of the quality of our model.  Nevertheless, our model \emph{is}
able to identify large parts of the the Meso-American area, the Baltic
area and the South Asian area.  (It also finds the Balkans, but since
these languages are all IE, we do not consider it a linguistic area in
this evaluation.)  While our model does find areas that match
Meso-American and North-west American areas, neither is represented in
its entirety (according to the definition of these areas given in
Section~\ref{sec:areas}).

Despite the difficulty humans have in assigning linguistic areas, In
Table~\ref{tab:clustering}, we explicitly compare the quality of the
areal clusters found on the IE subset.  We compare against the most
inclusive areal lists from Section~\ref{sec:areas} for IE: the Balkans
and the Baltic.  When there is overlap (eg., Romani appears in both
lists), we assigned it to the Balkans.

\begin{table}[t]
  \footnotesize
  \centering
  \begin{tabular}{|l|cccc|}
    \hline
    {\bf Model} & {\bf Rand} & {\bf F-Sc} & {\bf Edit} & {\bf NVI} \\
    \hline
    K-means     & $     0.9149$ & $        0.0735$ & $        0.1856$ & $        0.5889$ \\
    Pitman-Yor  & $     0.9637$ & $        0.1871$ & $        0.6364$ & $        0.7998$ \\
    Areal model & $\bf  0.9825$ & $\bf     0.2637$ & $\bf     0.8295$ & $\bf     0.9090$ \\
    \hline
  \end{tabular}
  \mycaption{Area identification scores for two baseline algorithms (K-means and Pitman-Yor clustering) that do not use hierarchical structure, and for the Areal model we have presented.  Higher is better and all differences are statistically significant at the 95\% level.}
  \label{tab:clustering}
\end{table}

We compare our model with a flat Pitman-Yor model that does not use
the hierarchy.  We also compare to a baseline $K$-means algorithm.
For $K$-means, we ran with $K \in \{5,10,15,\dots,80,85\}$ and chose
the value of $K$ for each metric that did best (giving an unfair
advantage).  Clustering performance is measured on the Indo-European
task according to the Rand Index, F-score, Normalized Edit Score
\cite{pantel-thesis} and Normalized Variation of Information
\cite{meila03clusterings}.  In these results, we see that the
Pitman-Yor process model dominates the $K$-means model and the Areal
model dominates the Pitman-Yor model.

\mysubsection{Identifying Areal Features} \label{sec:experiment-features}

Our second experiment is an analysis of the \emph{features} that tend
to be shared areally (as opposed to genetically).  For this
experiment, we make use of the whole-world version of the data, again
with known language family structure.  We initialize a Gibbs sampler
from the MAP configuration found in
Section~\ref{sec:experiment-areas}.  We run the sampler for 1000
iterations and take samples every ten steps.

From one particular sample, we can estimate a posterior distribution
over each $\pi_f$.  Due to conjugacy, we obtain a posterior
distribution of $\pi_f \by \Bet(1 + \sum_{n} 
Z_{n,f}, 1 + \sum_{n} [1-Z_{n,f}])$.  
The $1$s come from the prior.  From this Beta distribution, we can ask
the question: what is the probability that a value of $\pi_f$ drawn
from this distribution will have value $<0.5$?  If this value is high,
then the feature is likely to be a ``genetic feature''; if it is low,
then the feature is likely to be an ``areal feature.''  We average
these probabilities across all 100 samples.

\begin{table}[t]
\footnotesize
\centering
\begin{tabular}{|c|r|l|}
\hline
 {\bf p(gen)} & {\bf \#f} & {\bf Feature Category} \\
\hline
 .00 &	1&	Tea                                       \\
 .73 &	19&	Phonology                                 \\
 .73 &	9&	Lexicon                                   \\
 .74 &	4&	Nominal Categories / Numerals             \\
 .79 &	5&	Simple Clauses / Predication              \\
 .80 &	5&	Verbal Categories / Tense and Aspect      \\
 .87 &	8&	Nominal Syntax                            \\
 .87 &	8&	Simple Clauses / Simple Clauses           \\
 .91 &	12&	Nominal Categories / Articles and Pronouns\\
 .94 &	17&	Word Order                                \\
 .99 &	10&	Morphology                                \\
 .99 &	6&	Simple Clauses / Valence and Voice        \\
 .99 &	7&	Complex Sentences                         \\
 .99 &	7&	Nominal Categories / Gender and Number    \\
 .99 &	5&	Simple Clauses / Negation and Questions   \\
 1.0 &	1&	Other / Clicks                            \\
 1.0 &	2&	Verbal Categories / Suppletion            \\
 1.0 &	9&	Verbal Categories / Modality              \\
 1.0 &	4&	Nominal Categories / Case                 \\
\hline
\end{tabular}
\mycaption{Average probability of genetic for each feature
  category and the number of features in that category.}
\label{tab:areal-categories}
\end{table}

The features that are \emph{most} likely to be areal according to our
model are summaries in Table~\ref{tab:areal-categories}.  In this
table, we list the \emph{categories} to which each feature belongs,
together with the number of features in that category, and the
\emph{average} probability that a feature in that category is
genetically transmitted.  Apparently, the vast majority of features
are \emph{not} areal.

We can treat the results presented in Table~\ref{tab:areal-categories}
as a hierarchy of borrowability.  In doing so, we see that our
hierarchy agrees to a large degree with the hierarchies summarized in
Section~\ref{sec:features}.  Indeed, (aside from ``Tea'', which we
will ignore) the two most easily shared categories according to our
model are phonology and the lexicon; this is in total agreement with
the agreed state of affairs in linguistics.

Lower in our list, we see that noun-related categories tend to precede
their verb-related counterparts (nominal categories before verbal
categores, nominal syntax before complex sentences).  According to
Curnow~\shortcite{curnow01borrowability}, the most difficult features
to borrow are phonetics (for which we have no data), bound grammatical
forms (which appear low on our list), morphology (which is $99\%$
genetic, according to our model) and syntactic frames (which would
roughly correspond to ``complex sentences'', another item which is
$99\%$ genetic in our model).



\mysubsection{Genetic Reconstruction} \label{sec:experiment-tree}

In this section, we investigate whether the use of areal knowledge can
improve the automatic reconstruction of language family trees.  We use
Kingman's coalescent (see Section~\ref{sec:coalescent}) as a
probabilistic model of trees, endowed with a binomial mutation process
on the language features.

Our baseline model is to run the vanilla coalescent on the WALS data,
effective reproducing the results presented by Teh et
al. \shortcite{teh07coalescent}.  This method was already shown to
outperform competing hierarchical clustering algorithms such as
average-link agglomerative clustering (see, eg., Duda and
Hart~\shortcite{duda73book}) and the Bayesian Hierarchical Clustering
algorithm \cite{heller05bhc}.



We run the same experiment both on the IE subset of data
and on the whole-world subset.  We evaluate the results qualitatively,
by observing the trees found (on the IE subset) and
quantitatively (below).  For the qualitative analysis, we show the
subset of IE that does not contain Indic languages or
Iranian languages (just to keep the figures small).  The tree derived
from the original data is on the left in Figure~\ref{fig:ie-orig}, below:

\begin{figure*}[t]
\centering
\begin{tabular}{p{.45\textwidth}p{.45\textwidth}}
  \includegraphics[clip=true,height=0.45\textwidth,angle=90]{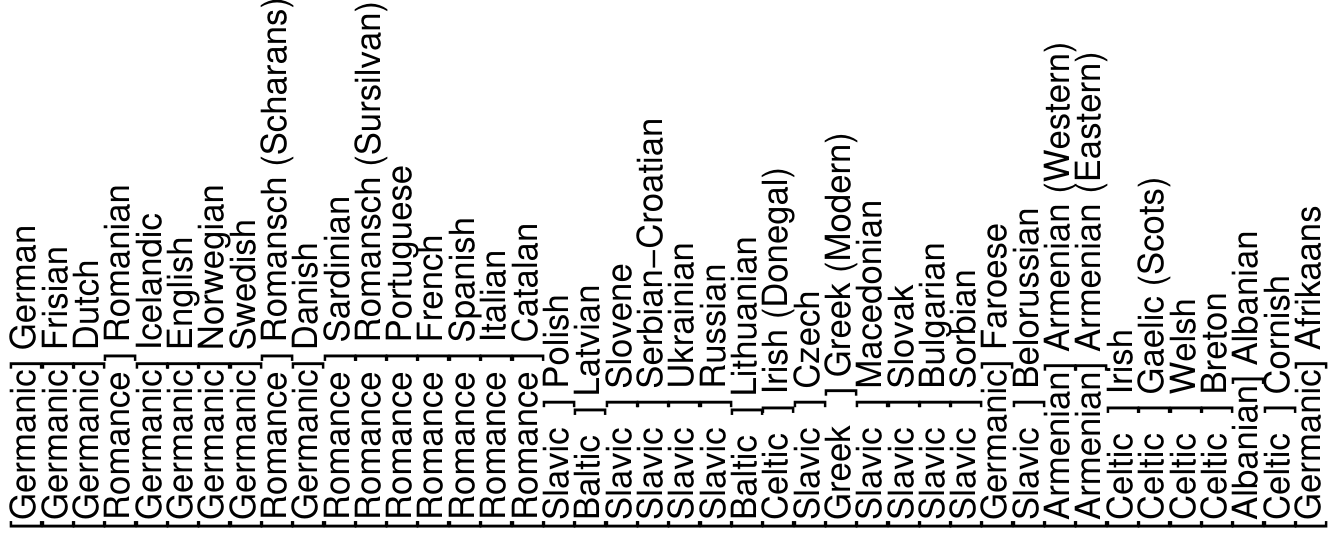}

\vspace{-0.3em}
  \includegraphics[clip=true,height=0.45\textwidth,width=1cm,angle=90]{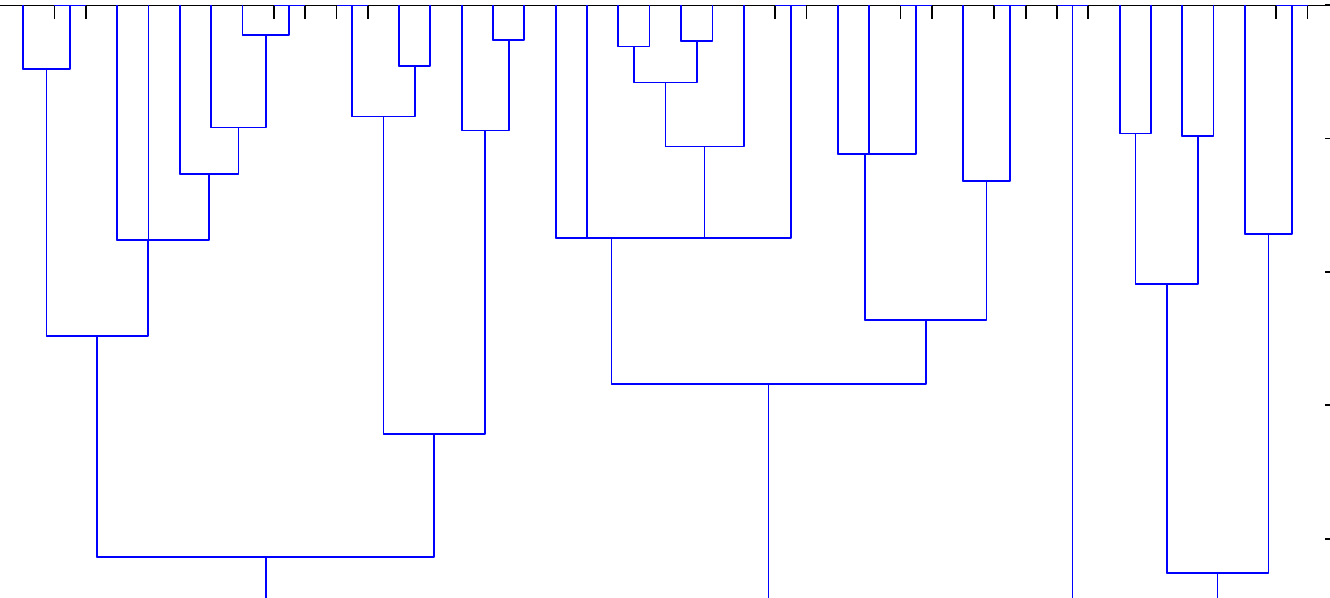}
&
  \includegraphics[clip=true,height=0.45\textwidth,angle=90]{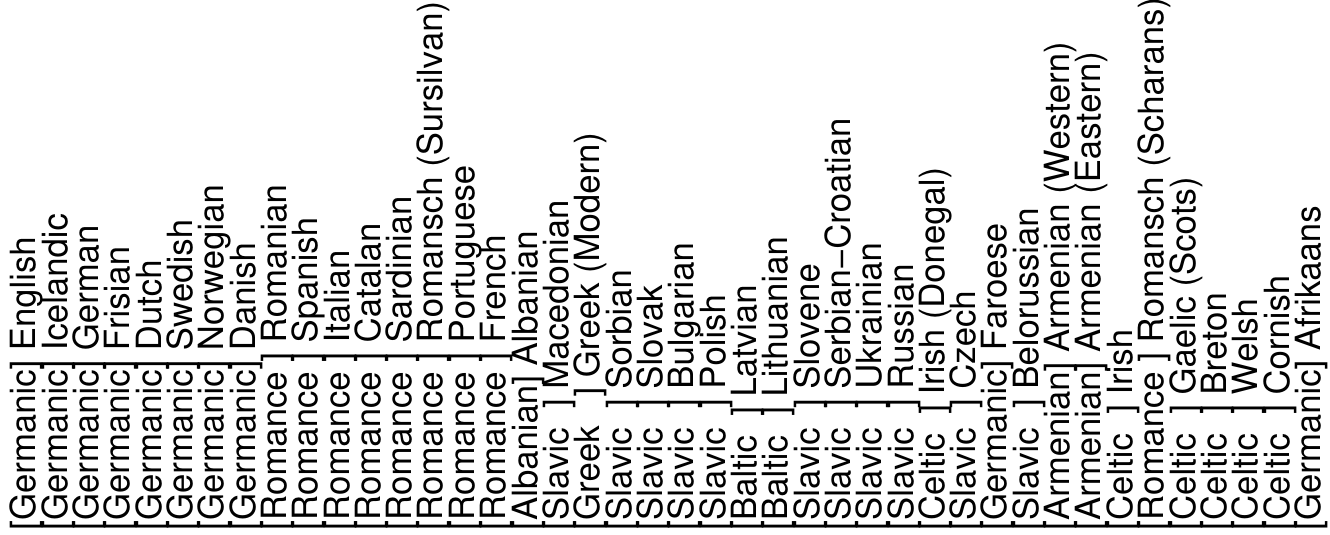}

\vspace{-0.3em}
  \includegraphics[clip=true,height=0.45\textwidth,width=1cm,angle=90]{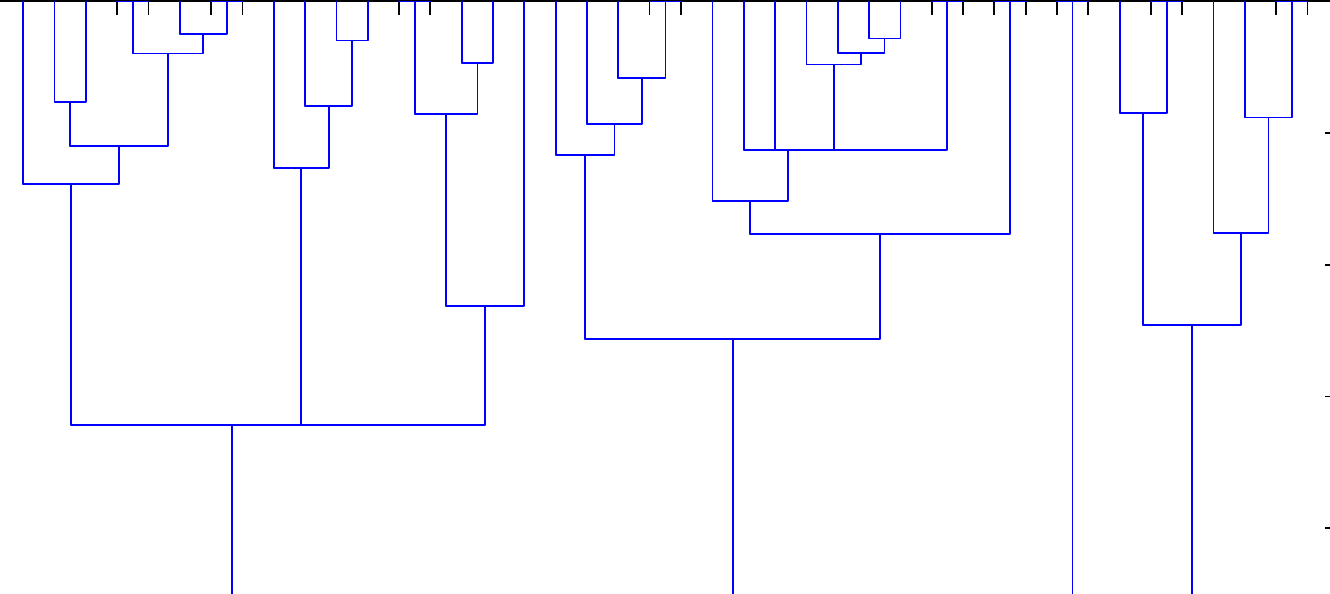}
\end{tabular}
  \mycaption{Genetic trees of IE languages.  (Left) with no areal
    knowledge; (Right) with areal model.}
  \label{fig:ie-orig}
\end{figure*}

\begin{table}[t]
  \footnotesize
  \centering
  {\bf Indo-European}
  \begin{tabular}{|l|c|c|}
    \hline
    {\bf Model} & {\bf Accuracy} & {\bf Log Prob} \\
    \hline
    Baseline    & $0.635$ ($\pm 0.007$) & $-0.583$ ($\pm 0.008$) \\
    Areal model & $\bf 0.689$ ($\pm 0.010$) & $\bf -0.526$ ($\pm 0.027$) \\
    \hline
  \end{tabular}
  {\bf World}
  \begin{tabular}{|l|c|c|}
    \hline
    {\bf Model} & {\bf Accuracy} & {\bf Log Prob} \\
    \hline
    Baseline    & $0.628$ ($\pm 0.001$) & $-0.654$ ($\pm 0.003$) \\
    Areal model & $\bf 0.635$ ($\pm 0.002$) & $\bf -0.565$ ($\pm 0.011$) \\
    \hline
  \end{tabular}
  \mycaption{Prediction accuracies and log probabilities for
    IE (top) and the world (bottom).}
  \label{tab:reconstruct-results}
\end{table}

\noindent
The tree based on areal information is on the right in
Figure~\ref{fig:ie-orig}, below.  As we can see, the use of areal
information qualitatively improves the structure of the tree.  Where
the original tree had a number of errors with respect to Romance and
Germanic languages, these are sorted out in the areally-aware tree.
Moreover, Greek now appears in a more appropriate part of the tree and
English appears on a branch that is further out from the Norse
languages.

We perform two varieties of quantitative analysis.  In the first, we
attempt to predict unknown feature values.  In particular, we
\emph{hide} an addition $10\%$ of the feature values in the WALS data
and fit a model to the remaining $90\%$.  We then use that model to
predict the hidden $10\%$.  The baseline model is to make predictions
according to the family tree.  The augmented model is to make
predictions according to the family tree \emph{for those features
  identified as genetic} and according to the linguistic area
\emph{for those features identified as areal.}  For both settings, we
compute both the absolute accuracy as well as the log probability of
the hidden data under the model (the latter is less noisy).  We repeat
this experiment $10$ times with a different random $10\%$ hidden.  The
results are shown in Table~\ref{tab:reconstruct-results}, below.  The
differences are not large, but are outside one standard deviation.

For the second quantitative analysis, we use present purity scores
\cite{heller05bhc}, subtree scores (the number of interior nodes with
pure leaf labels, normalized) and leave-one-out log accuracies (all
scores are between 0 and 1, and higher scores are better).  These
scores are computed against both language family and language genus as
the ``classes.''  The results are in Table~\ref{tab:purity-results},
below.  As we can see, the results are generally in favor of the Areal
model (LOO Acc on IE versus Genus nonwithstanding), depending on
the evaluation metric.

\begin{table}[t]
  \footnotesize
  \centering
  {\bf Indo-European versus Genus}
  \begin{tabular}{|l|c|c|c|}
    \hline
    {\bf Model} & {\bf Purity} & {\bf Subtree} & {\bf LOO Acc} \\
    \hline
    Baseline    & $    0.6078$ & $    0.5065$ & $\bf 0.3218$ \\
    Areal model & $\bf 0.6494$ & $\bf 0.5455$ & $    0.2528$ \\
    \hline
  \end{tabular}

  {\bf World versus Genus}
  \begin{tabular}{|l|c|c|c|}
    \hline
    {\bf Model} & {\bf Purity} & {\bf Subtree} & {\bf LOO Acc} \\
    \hline
    Baseline    & $    0.3599$ & $    0.2253$ & $    0.7747$ \\
    Areal model & $\bf 0.4001$ & $\bf 0.2450$ & $\bf 0.7982$ \\
    \hline
  \end{tabular}

  {\bf World versus Family}
  \begin{tabular}{|l|c|c|c|}
    \hline
    {\bf Model} & {\bf Purity} & {\bf Subtree} & {\bf LOO Acc} \\
    \hline
    Baseline    & $    0.4163$ & $    0.3280$ & $    0.4842$ \\
    Areal model & $\bf 0.5143$ & $\bf 0.3318$ & $\bf 0.5198$ \\
    \hline
  \end{tabular}
  \mycaption{Scores for IE as compared against genus
    (top); for world against genus (mid)
    and against family (low).}
  \label{tab:purity-results}
\end{table}

\mysubsection{Effect of Radius} \label{sec:radius}

\begin{table}[t]
  \footnotesize
  \centering
  \begin{tabular}{|l|c|c|c|}
    \hline
    {\bf Radius} & {\bf Purity} & {\bf Subtree} & {\bf LOO Acc} \\
    \hline
     $125$ & $0.6237$ &   $0.4855$ &   $0.2013$ \\
     $250$ & $0.6457$ &   $0.5325$ &   $0.2299$ \\
     $500$ & $0.6483$ &   $\bf 0.5455$ &   $0.2413$ \\
    $1000$ & $\bf 0.6494$ &   $\bf 0.5455$ &   $0.2528$ \\
    $2000$ & $0.6464$ &   $0.4935$ &   $0.3218$ \\
    $4000$ & $0.6342$ &   $0.4156$ &   $\bf 0.4138$ \\
    \hline
  \end{tabular}
  \mycaption{Scores for IE vs genus at varying radii.}
  \label{tab:radii}
\end{table}

Finally, we evaluate the effect of the radius hyperparameter on
performance.  Table~\ref{tab:radii} shows performance for models built
with varying radii.  As can be seen by purity and subtree scores,
there is a ``sweet spot'' around $500$ to $1000$ kilometers where the
model seems optimal.  LOO (strangely) seems to continue to
improve as we allow areas to grow arbitrarily large.  This is perhaps
overfitting.  Nevertheless, performance is robust for a range of
radii.

\mysection{Discussion} \label{sec:discussion}

We presented a model that is able to recover well-known linguistic
areas.  Using this areas, we have shown improvement in the ability to
recover phylogenetic trees of languages.  It is important to note that
despite our successes, there is much at our model does not account
for: borrowing is known to be assymetric; contact is temporal;
borrowing must obey univeral implications.  Despite the failure of our
model to account for these issues, however, it appears largely
successful.  Moreover, like any ``data mining'' expedition, our model
suggests new linguistic areas (particularly in the ``whole world''
experiments) that deserve consideration.

\vspace{-2mm}
\subsection*{Acknowledgments}
\vspace{-1mm}

Deep thanks to Lyle Campbell, Yee Whye Teh and Eric Xing for
discussions; comments from the three anonymous reviewers were very
helpful.  This work was partially supported by NSF grant IIS0712764.

\bibliographystyle{naaclhlt2009}
\bibliography{bibfile}

\end{document}